%% file: acl_latex.tex
\documentclass[11pt]{article}
\usepackage{newtxtext}

\usepackage[final]{acl}

\usepackage[normalem]{ulem}
\useunder{\uline}{\ul}{}
\usepackage{times}
\usepackage{latexsym}
\usepackage{amsmath}
\usepackage{booktabs}
\usepackage{hyperref}
\usepackage{url}
\usepackage{booktabs}
\usepackage{graphicx}
\usepackage{xspace}
\usepackage{amsmath}
\usepackage{amssymb}
\usepackage{multirow}
\usepackage{diagbox}
\usepackage{array}
\usepackage[table]{xcolor}
\usepackage[most]{tcolorbox}
\usepackage{colortbl}    
\usepackage{makecell}
\usepackage{tabularx}
\usepackage{enumitem}
\usepackage{tipa}
\usepackage{cleveref}
\usepackage{subcaption}
\usepackage{caption}
\usepackage{algorithm}
\usepackage{algorithmicx}
\usepackage{algpseudocode}
\usepackage[T1]{fontenc}
\usepackage[utf8]{inputenc}

\usepackage{microtype}

\usepackage{inconsolata}

\usepackage{graphicx}
\newcommand{\Ours}{\textsc{GeoDe}\xspace}
\title{Purging the Gray Zone: Latent-Geometric Denoising for Precise Knowledge Boundary Awareness}

\author{
  Hao An\thanks{Equal contribution.},
  Yibin Lou\footnotemark[1],
  Jiayi Guo,
  Yang Xu\thanks{{Correspondence:} \href{mailto:xuyang@sustech.edu.cn}{xuyang@sustech.edu.cn}.} \\
  Computational Linguistics and Consciousness Sciences Lab \\Southern University of Science and Technology
}

\begin{document}
\maketitle
\begin{abstract}
Large language models (LLMs) often exhibit hallucinations due to their inability to accurately perceive their own knowledge boundaries. Existing abstention fine-tuning methods typically partition datasets directly based on response accuracy, causing models to suffer from severe label noise near the decision boundaries and consequently exhibit high rates of abstentions or hallucinations. This paper adopts a latent space representation perspective, revealing a ``gray zone'' near the decision hyperplane where internal belief ambiguity constitutes the core performance bottleneck. Based on this insight, we propose the \emph{\Ours} (\textbf{Geo}metric \textbf{De}noising) framework for abstention fine-tuning. This method constructs a truth hyperplane using linear probes and performs ``geometric denoising'' by employing geometric distance as a confidence signal for abstention decisions. This approach filters out ambiguous boundary samples while retaining high-fidelity signals for fine-tuning. Experiments across multiple models (Llama3, Qwen3) and benchmark datasets (TriviaQA, NQ, SciQ, SimpleQA) demonstrate that \Ours significantly enhances model truthfulness and demonstrates strong generalization in out-of-distribution (OOD) scenarios. Code is available at \href{https://github.com/Notbesidemoon/GeoDe}{https://github.com/Notbesidemoon/GeoDe}.
\end{abstract}

\section{Introduction}
Large Language Models (LLMs) have demonstrated outstanding performance across various natural language processing tasks~\citep{llama3,yang2025qwen3technicalreport,zhu-etal-2024-zero}. However, it is universally acknowledged that LLMs exhibit hallucination—that is, generating responses that are factually inaccurate or fabricating answers~\citep{zhang2023siren}. This issue underscores the urgent need to develop effective hallucination detection and mitigation methods.

\begin{figure}[!htbp]
    \centering
\includegraphics[width=0.45\textwidth]{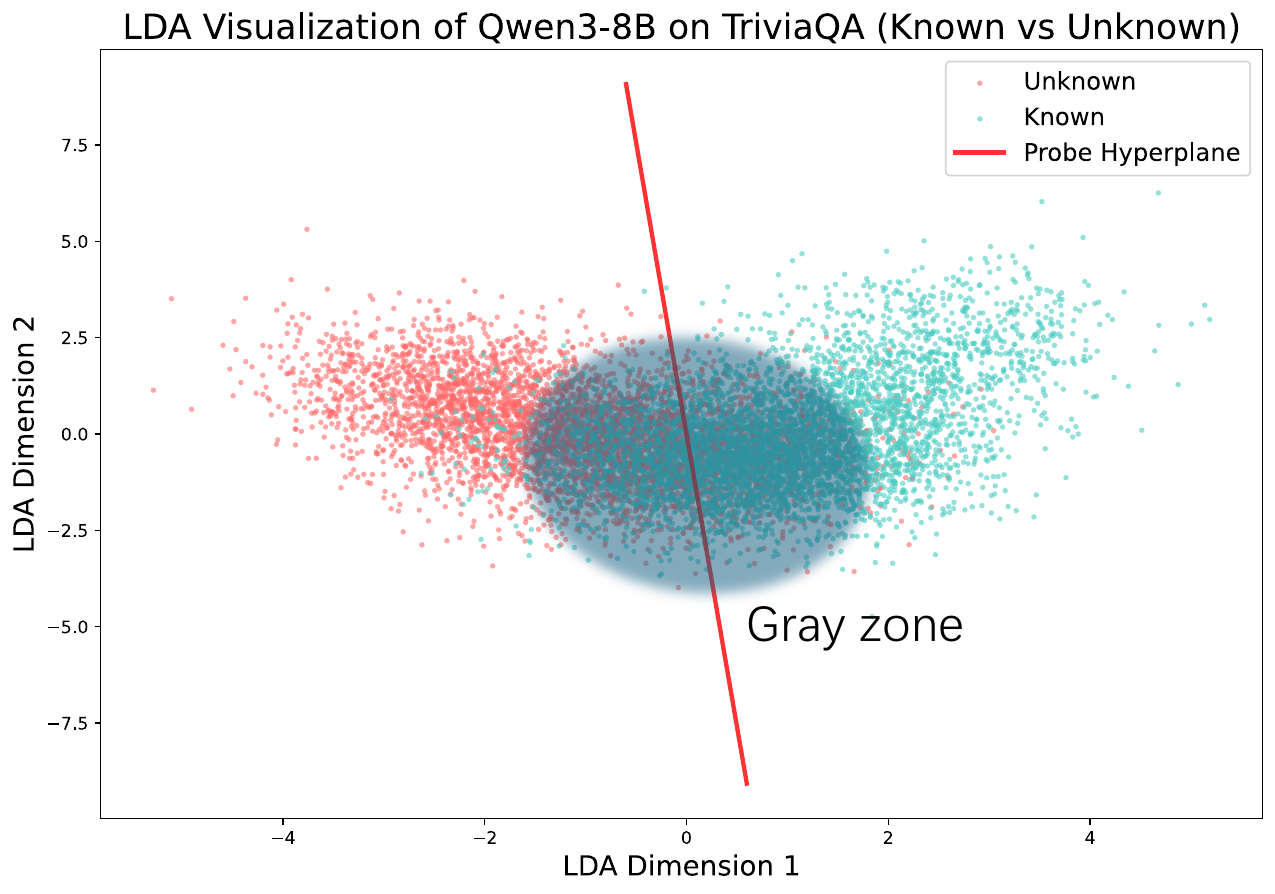}
    \caption{Visualization of the hidden states of questions that are known and unknown to the model. ``Gray zone'' refers to the overlapping area.}
    \label{fig:lda_acc}
\end{figure}

% Abstention fine-tuning can mitigate hallucination
% However, it will over-abstain and 
One practical approach to mitigating hallucination is to fine-tune LLMs to answer known questions while abstaining from those beyond their knowledge scope~\citep{abstention_survey, li-etal-2025-knowledge-boundary}. Specifically, these methods typically classify training data into ``known'' and ``unknown'' questions based on the correctness of model responses, training models to answer the known set while replying ``I don't know'' to the unknown set~\citep{R-Tuning, Can_AI_assistants_know}. To reduce reliance on ground truth labels, alternative approaches utilize uncertainty metrics like semantic entropy to partition training data into known and unknown questions~\citep{SE_Tuning, ualign}.

These methods have shown notable effectiveness, yet they often fail when internal confidence misaligns with external correctness. Relying on response-level accuracy to partition ``known'' and ``unknown'' sets introduces significant label noise—such as ``lucky guesses'' or formatting-driven failures. Training on these noisy heuristics forces the model to learn an inconsistent, contradictory decision boundary, ultimately leading to over-refusal or persistent hallucinations. We employ a \emph{probing} method to analyze such cases. As shown in \Cref{fig:lda_acc}, we visualize the known and unknown sets, with the red line representing the learned probe hyperplane. The two sets exhibit significant overlap in the central region, whereas samples farther from the hyperplane show clean separation. This indicates that a significant portion of current abstention fine-tuning data contains noise.
Therefore, a potential improvement is to discard noisy boundary samples and retain only clean, high-confidence data, so that the LLM can learn to distinguish known from unknown cases more effectively.

To this end, we propose the \textbf{Geo}metric \textbf{De}noising (\Ours) framework for abstention fine-tuning, inspired by the linear representation hypothesis. The key intuition is that a hidden state far from the probe hyperplane indicates high confidence, making the abstention decision straightforward, whereas a state near the hyperplane reflects ambiguity, making the decision unreliable. Guided by this principle, we select samples with high probe confidence (distant from the probe hyperplane) and discard those with low probe confidence (close to the probe hyperplane).

% \begin{figure}[!htbp]

%     \centering
%     \includegraphics[width=0.45\textwidth]{figure/S1-Intro-knowledge.pdf}
%     \label{fig:knowledge_boundary}
%     \caption{Illustration of over-abstention and persistent hallucination.}
    
% \end{figure}

Our main contributions are as follows:
\begin{enumerate}[leftmargin=*]
\item \textbf{Internal Representation Perspective}: We offer a novel diagnostic perspective on abstention fine-tuning by analyzing the latent space of LLMs. Our analysis reveals that suboptimal performance frequently stems from a ``grey zone'' near the latent decision boundary, where ambiguous representations introduce significant label noise.

\item \textbf{Latent-Guided Denoised Dataset Curation}: We propose \Ours, a framework that leverages internal probes to curate high-quality fine-tuning datasets. By using geometric distance from the truthfulness hyperplane as a confidence metric, \Ours purges ambiguous boundary samples. This geometric denoising ensures the model trains on linearly separable signals, leading to sharper knowledge boundaries.

\item \textbf{Empirical Superiority}: Extensive experiments across multiple architectures and benchmarks demonstrate that \Ours significantly outperforms baselines. Our method also shows superior generalization in out-of-distribution (OOD) and abstention tasks.
\end{enumerate}

\section{Related Work}

\subsection{Hidden States of LLMs}
Recent work suggests there is a ``truthfulness’’ direction in latent space~\citep{geometry_of_truth,probing_llm_lying}. \citet{liu-etal-2024-universal} suggest there is a universal truthfulness hyperplane within LLMs that generalizes across tasks, domains, and in-domain settings. Some work probes the last token of a question to predict whether the model can answer it correctly without generating any tokens~\citep{slobodkin-etal-2023-curious, 10.1145/3637528.3671796, estimating_without_generate}. To more effectively distinguish facts from errors, some work designs more complex features to train truthfulness probes and utilizes information from model-generated answers~\citep{exact_probe, li-etal-2025-hd, icr-probe}. Truthfulness vectors can also be employed for hallucination mitigation via steering~\citep{ji-etal-2025-calibrating, zhang-etal-2024-truthx}. Recent works suggest that models’ own internal judgments often lead to better overall factuality~\citep{curious_hidden, liang2024learningtrustfeelingsleveraging}. In this work, we employ the truthfulness hyperplane as an internal confidence classifier to guide abstention fine-tuning.

\subsection{Abstention Fine-tuning} Abstention fine-tuning is a technique that teaches the model to abstain from answering questions whose answers it does not know, while maintaining accuracy on known questions~\citep{abstention_survey}. \citet{R-Tuning,SE_Tuning,Can_AI_assistants_know} construct an abstention-aware dataset based on whether the model can answer correctly, defining this as the model's knowledge boundary. Then they fine-tune the model to refuse to answer questions beyond its knowledge boundary while responding to those within it. \citet{xurejection,Can_AI_assistants_know,brahman2024the} use Direct Preference Optimization (DPO)~\citep{DPO} to train models to admit uncertainty when encountering unknown questions rather than outputting incorrect answers. \citet{li2025refine} employ adaptive contrastive learning to optimize LLMs' abstention preferences.
\citet{rej_token, idk_token} incorporate a dedicated ``rejection'' token into the model's vocabulary and formulate an objective function that redistributes probability mass toward this token when the model is uncertain. \citet{zheng2025enhancing, FSCR} train models to output binary confidence labels (``sure'' vs. ``unsure'') after generating an answer as a proxy for abstention, enabling them to reject low‑confidence answers. Abstention fine-tuning may result in models being overly conservative (over-abstaining) or overly aggressive (hallucinating)~\citep{Can_AI_assistants_know, zhu-etal-2025grait}. In this work, we construct fine-tuning datasets based on the model's internal beliefs to mitigate issues of over-rejection and over-hallucination.

\section{Method}
\begin{figure*}[htbp]
    \centering
    \includegraphics[width=0.95\linewidth]{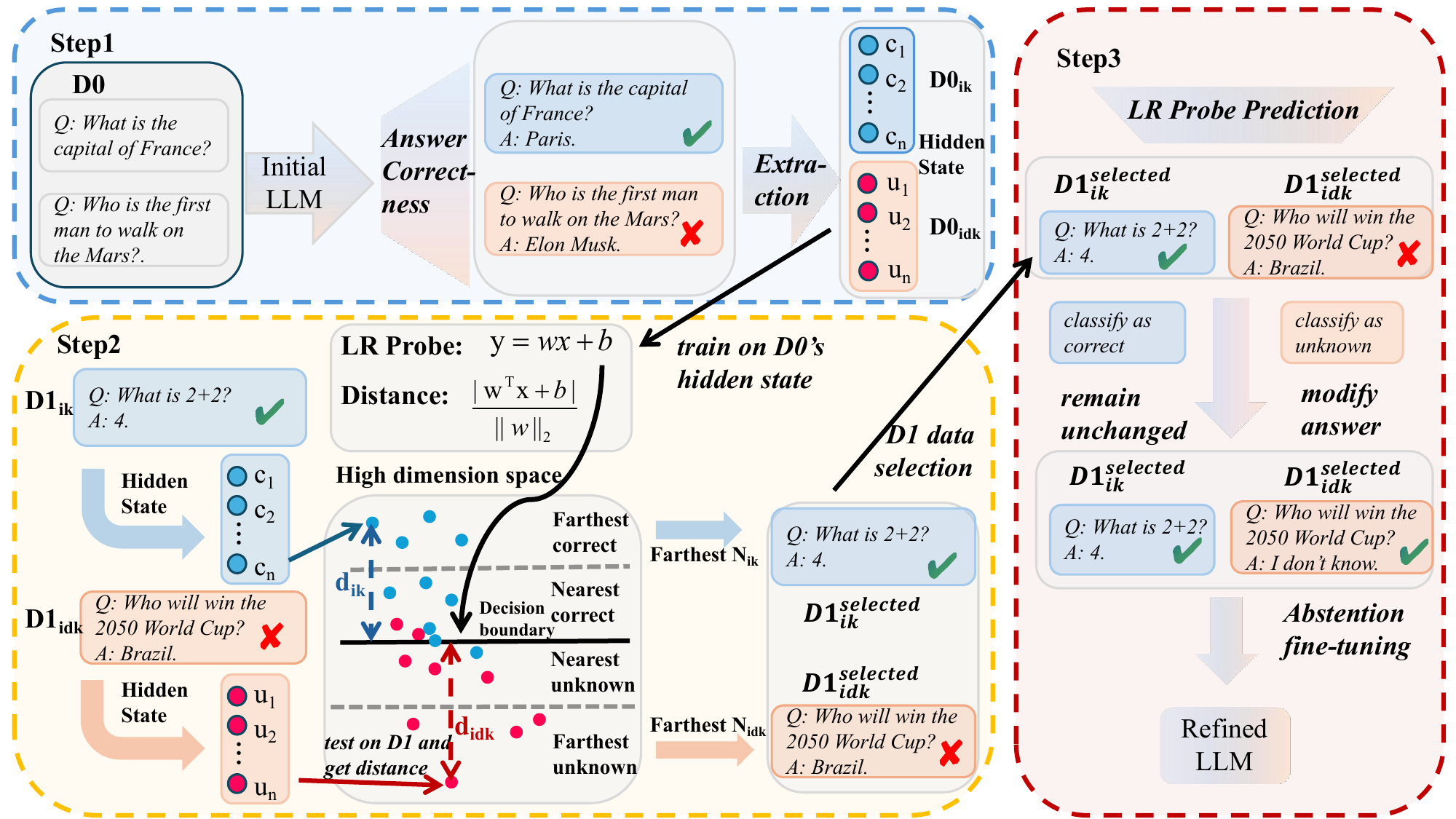}
    \caption{Overview of our method. \Ours contains three steps: (1) Identify the knowledge boundary by dividing the source data into two subsets ($\text{D0}_{\text{ik}}$ and $\text{D0}_{\text{idk}}$), and train a probe using these two subsets. (2) Calculate the distance from test data D1 to the hyperplane learned by the probe, and select the subset of samples that are farthest from the hyperplane $D1_{\text{ik}}^{\text{selected}}$ and $D1_{\text{idk}}^{\text{selected}}$. (3) Adjust the target answers based on the probe's prediction results (retain correct answers for $D1_{\text{ik}}^{\text{selected}}$ and replace incorrect answers with ``I don't know'' for $D1_{\text{idk}}^{\text{selected}}$), then conduct abstention fine-tuning on the selected subset.}
    \label{fig:overview}
\end{figure*}
Our work aims to develop a hidden-state-based approach to enhance the model’s awareness of its own knowledge boundaries. The main steps are to partition the training data into known/unknown subsets, train a hidden-state probe to quantify the model’s confidence in its responses, and finally perform targeted abstention fine-tuning on the curated samples to teach the model to abstain from answering questions outside its knowledge scope.

\subsection{Identify the Knowledge Boundary}
First, the base LLM is prompted to answer every question in a source training dataset ($D0$). This serves as a diagnostic phase to examine what knowledge the model has internalized. The model's responses are then split into two distinct subsets based on whether each response is correct.

\noindent{\textbf{Known Knowledge}} ($D0_{\text{ik}}$): These are samples for which the LLM provided a correct response. The ``ik'' stands for ``I know''. These question-answer pairs are retained in their original form to reinforce the retention of correct information during fine-tuning.

\noindent{\textbf{Unknown Knowledge }}($D0_{\text{idk}}$): These are samples for which the LLM provided an incorrect response.
For these cases, the original (incorrect) answer is discarded and replaced with a refusal message---specifically ``I don't know'' in abstention fine-tuning (hence, ``idk'' for short).

\subsection{Curating Denoised Dataset}
We train the probe model using $D0_{\text{ik}}$ and $D0_{\text{idk}}$ from the previous step. Specifically, we use the hidden state representation of the question as the feature $\mathbf{x}=f_\text{LLM}(q)$, and the correctness of the statement as a binary label $y=\mathbb{I}(q\Rightarrow a)\in\{0,1\}$, to train a logistic regression probe, whose formula is:
\begin{equation}
    f_\text{probe}(\mathbf{x})=\sigma(\mathbf{w}^\top \mathbf{x}+b),
\end{equation}
where $\sigma$ is the sigmoid function, and $w$ denotes the linear weight and $b$ is the bias term. Next, we define the confidence of LLM answering question $q$ with $a$ by measuring the distance from $\mathbf{x}$ to the learned hyperplane $(\mathbf{w};b)$:
\begin{equation}
    d(\mathbf{x}) = \frac{\left| \mathbf{w}^{\top} \mathbf{x} + b \right|}{\left\lVert \mathbf{w} \right\rVert_{2}}
    \label{eq:distance}
\end{equation}

$d(\mathbf{x})$ has a clear geometric meaning that reflects the model's confidence in answering the given question. $d(\mathbf{x}) > 0$ means the model believes it can answer correctly, while $d(\mathbf{x}) < 0$ means the opposite, and a larger $|d(\mathbf{x})|$ value indicates stronger confidence.
It follows naturally that we can reformulate the goal of abstention fine-tuning as teaching the model to reject questions for which $d(\mathbf{x}) < 0$ and to answer those for which $d(\mathbf{x})>0$.
% Therefore, we hypothesize that questions with small \( |d(x)| \) make it difficult for the model to decide whether to answer or abstain, while questions with large \( |d(x)| \) are easier to judge. 
Accordingly, we partition the data into subsets of varying difficulty based on the magnitude of $|d(\mathbf{x})|$. Instead of using all available data, we select only samples farthest from the decision boundary, i.e., those with large $|d(\mathbf{x})|$, retaining the top $X\%$ of samples for the fine-tuning task.

The two approaches for extracting the hidden state from LLMs, $\mathbf{x}=f_\text{LLM}(q)$, are as follows:
\paragraph{Hidden state of the question (TBG):}
We directly feed the question string $q$ into the model and extract the hidden state of the last token from the final layer, i.e., the \textbf{token before generation} (TBG)—the last token of the question, immediately before the model begins generating its response.
\paragraph{Hidden state of the answer (SLT):}
We first generate the model's answer $a$ via few-shot learning and greedy decoding (with temperature set to 0), then feed the concatenated sequence $q\oplus a$ back into the model to retrieve the hidden state of the last token from the final layer, i.e., the \textbf{second last token} (SLT); this token captures the contextual representation of the entire question-answer sequence right before the end-of-sequence token.

\subsection{Abstention Fine-tuning on Subset}
% The purpose of abstention fine-tuning is to enable the model to distinguish between known and unknown questions, thereby choosing to answer or abstain. 
Given a dataset ${D}^\text{selected}$ that consists of a known question set ${D}_\text{ik}^\text{selected}$ and an unknown question set ${D}_\text{idk}^\text{selected}$, we modify the ground truth of ${D}_\text{idk}^\text{selected}$ as ``I don't know'' and keep the ground truth of ${D}_\text{ik}^\text{selected}$ as the correct answer. Then we employ supervised fine-tuning with the cross-entropy loss:
\begin{equation}
    \mathcal{L}_{(p_\theta)} = - \sum_{q \in D^\text{selected}} \sum_{t = 1}^{\lvert y^{(q)} \rvert} \log p_\theta \bigl( y^{(q)}_t \mid \text{I}, \mathbf{q}, \mathbf{y}^{(q)}_{t-1} \bigr),
    \label{eq:ce_loss}
\end{equation}
in which $p_\theta(y_t^{(q)})$ is the model's predicted next-token probability distribution given the instruction (I), question (q), and the first $t-1$ tokens of the ground truth $\mathbf{y}^{(q)}_{t-1}$.

\input{table/algotithm}

\section{Experiments}
\subsection{Experimental Setting}

\paragraph{Datasets} We assess the effectiveness of \Ours on four open-ended question-answering tasks: \textbf{TriviaQA}~\citep{triviaqa} which contains general knowledge QA pairs; \textbf{Natural Questions} (NQ)~\citep{NQ_open} which contains questions from users' queries to search engines; \textbf{SciQ}~\citep{SciQ} which contains science exam questions across multiple disciplines; and \textbf{SimpleQA}~\citep{wei2024measuringshortformfactualitylarge}, collected adversarially to challenge GPT-4, consisting of short-answer factual questions with single, indisputable answers to test whether the model truly ``knows what it knows''. We use 10K samples from the TriviaQA training set for probe training, with the rest used for SFT. The validation split of TriviaQA is used for the in-domain (ID) test. We use NQ, SciQ, and SimpleQA for out-of-distribution (OOD) tests. Detailed evaluation dataset information is provided in Appendix~\ref{dataset_detail}.

\paragraph{Baselines}
We compare \Ours with the following existing methods for abstention fine-tuning.

\begin{enumerate}[leftmargin=1.5em, itemsep=0em, topsep=0em]
    \item \textbf{IDK~\citep{Can_AI_assistants_know}} directly prompts the model to abstain from uncertain questions.

    \item \textbf{Uncertainty~\citep{two_stage_prompt}} first prompts the model to answer questions as accurately as possible, then prompts the model to output binary uncertainty (sure or unsure).

    \item \textbf{R-Tuning~\citep{R-Tuning}} randomly selects a set of questions, categorizes them as known or unknown based on the model's accuracy on each question, changes the ground truth for unknown questions to ``I don't know,'' and then performs supervised fine-tuning. R-Tuning does not apply any additional filtering beyond this accuracy-based categorization. \textbf{R-Tuning-01} (used in ablations) is a stricter variant that first samples 10 answers per question and retains only those where all answers are either correct or incorrect, thereby filtering out ambiguous questions before fine-tuning.

    \item \textbf{Probe-Tuning~\citep{curious_hidden}}. The workflow of Probe-Tuning is identical to that of R-Tuning, with the key difference being that Probe-Tuning utilizes the prediction results from truthfulness probes as the criteria for classifying questions into known and unknown categories.
\end{enumerate}

\paragraph{Evaluation}
\begin{table}[htbp!]
  \centering

\input{table/Abstention_confusion_matrix}

\caption{Abstention confusion matrix. R denotes the answer of the refined (fine-tuned) model. GT denotes the initial model. ``Known'' denotes that the initial model answered correctly, while ``unknown'' indicates that the initial model answered wrongly.}
\label{tab:abstention_confusion_matrix}
\end{table}
For each test question, we classify the response as correct, wrong, or abstention. We use Llama3.1-8b-instruct~\citep{llama3} as the judge to evaluate the correctness of answers generated by LLMs with 6-shot prompting (inter-judge consistency analysis is provided in Appendix~\ref{appendix:judge_robustness}).  A response containing ``I don't know'' is counted as an abstention. To measure the performance of abstention fine-tuning methods, we adopt three metrics, each reflecting unique aspects of performance, based on the widely used abstention confusion matrix in Table~\ref{tab:abstention_confusion_matrix}. We use in-context learning (\textbf{ICL}) as the initial model.

\textbf{Helpfulness ($\text{F1}_\text{ans}$)}:
For known questions, we calculate $\text{F1}_\text{ans}$~\citep{kim-etal-2024-aligning} as the harmonic mean of answerable recall ($\frac{N_1}{N_1+N_2+N_3}$) and answerable precision ($\frac{N_1}{N_1+N_2+N_4}$).

\textbf{Truthfulness ($\text{F1}_\text{abs}$)}:
For unknown questions, we calculate $\text{F1}_\text{abs}$~\citep{kim-etal-2024-aligning} as the harmonic mean of unanswerable recall ($\frac{N_5}{N_4+N_5}$) and unanswerable precision ($\frac{N_5}{N_3+N_5}$).

\textbf{Reliability ($\text{F1}_\text{rel}$)}:
Existing studies indicate that enhancing helpfulness leads to a decline in factuality~\citep{xurejection, FSCR}. Therefore, we calculate the harmonic mean of metrics $\text{F1}_\text{ans}$ and $\text{F1}_\text{abs}$ as a reliability metric for comprehensive evaluation~\citep{FSCR}.

\paragraph{Implementation Details} 
In this work, we choose Llama3-8B-Instruct~\citep{llama3} and Qwen3-8B~\citep{yang2025qwen3technicalreport} as the initial models. We conduct experiments using SFT. We set $X = 25\%$. We employ a logistic regression probe with L2 regularization. We use the Swift\footnote{\url{https://github.com/modelscope/ms-swift}} framework to conduct fine-tuning using the AdamW optimizer, training for 3 epochs, with a learning rate of 1e-5 and a batch size of 16. We use grid search to select the optimal hyperparameters. All experiments are implemented on 4 Nvidia L40-48GB GPUs. During inference, we utilize the vLLM framework\footnote{\url{https://github.com/vllm-project/vllm}} to accelerate the process and employ a greedy search strategy to generate responses. Results are averaged over three different random seeds.
% Hyperparameters and additional configurations are detailed in Appendix~\ref{hyperparameter}.
\subsection{Main Results}
\begin{table*}[!htbp]

    \centering
    \input{table/main_reults}
    \caption{Performance on in-domain and out-of-domain question answering benchmarks. All results are multiplied by 100. The best result is bolded. The second best is underlined.}
    \label{tab:main_reslts}
\end{table*}
We show the main experimental results of \Ours and all baseline methods in Table~\ref{tab:main_reslts}. The key observations from our experiments are:
\paragraph{Effectiveness of \Ours}
The fundamental limitation of existing baselines lies in their susceptibility to the ``gray zone''—the latent region where internal belief is ambiguous and misaligned with external correctness. As illustrated in Figure~\ref{fig:lda_acc}, partitioning data based solely on response accuracy results in significant overlap between known and unknown representations. By utilizing the geometric distance from the truthfulness hyperplane, \Ours effectively purges this noise. Our results demonstrate that this denoising process allows the model to fine-tune on pure representations of its knowledge boundary. Consequently, \Ours consistently outperforms all baselines in the reliability metric $\text{F1}_\text{rel}$ across both Llama3-8B-Instruct and Qwen3-8B. For instance, on the TriviaQA in-domain task, \Ours (TBG) achieves a top $\text{F1}_\text{rel}$ of 77.1 for Llama3 and 77.6 for Qwen, representing a substantial improvement over traditional R-Tuning and Probe-Tuning. \Ours also exhibits exceptional OOD generalization. On NQ, SciQ and SimpleQA, \Ours consistently maintains high $\text{F1}_\text{rel}$ scores.

\paragraph{Latent Representations: TBG vs. SLT} TBG and SLT exhibit comparable performance across benchmarks, with no single strategy consistently outperforming the other. While their absolute gains are similar, they reflect cognitive states at different stages: TBG captures the model's initial confidence prior to output, whereas SLT incorporates the semantic context of the generated response. This result underscores the robustness of the \Ours framework regarding representation positioning. It demonstrates that geometric denoising effectively identifies and reduces noise regardless of the specific extraction point, confirming the universal utility of latent-geometric distance.

\subsection{Evaluation on RAG Setting}
\begin{table*}[!htbp]
  \begin{center} 
  \input{table/rag}
  \end{center}
  \caption{RAG results. Acc. is accuracy. Hallu. is hallucination rate.}
\label{tab:rag}
\end{table*}
\paragraph{Experimental Setting}
We evaluate the performance of the abstention methods in the Retrieval Augmented Generation (RAG) scenario. We use the RAG-Bench~\citep{RAG_bench} dataset, which includes two settings: (1)~\textbf{Golden}: golden retrieval, which contains contexts that include correct answers. (2)~\textbf{Golden \& RRN}: golden retrieval with relevant retrieval noise, which includes golden retrieval and context relevant to the question statement but lacks the correct answers. We feed the context alongside the question into the model.

Note that our RAG experiments deviate from the standard abstention-in-RAG formulation studied by prior works~\citep{joren2025sufficient, filice2025generate}, where ``unknown'' is typically assigned when the retrieved context does not contain the answer. In our setting, we define a question as ``unknown'' if the model \emph{fails to produce the correct answer}, even when the context contains it. This allows us to probe whether \Ours can transfer from mitigating memorization hallucinations to mitigating reading-comprehension hallucinations, covering cases where the model's extraction or reasoning ability is the limiting factor rather than retrieval quality.

\paragraph{Experimental Results}
Table~\ref{tab:rag} shows the performance in the RAG setting. Similar to the main experimental results, IDK and R-Tuning perform worse than abstention fine-tuning based on probing. Overall, our method achieves the best overall performance among all baselines. Within the SLT-based setting, our method not only achieves superior accuracy but also yields a lower hallucination rate compared to Probe-Tuning. This dual improvement highlights the efficacy of geometric denoising in establishing more reliable knowledge boundaries. Our method maintains the highest reliability ($\text{F1}_\text{rel}$) even under noisy retrieval scenarios, demonstrating its robustness in practical applications.

\section{Analysis}

\subsection{Ablation Studies}
\begin{figure}
    \centering
    \includegraphics[width=0.9\linewidth]{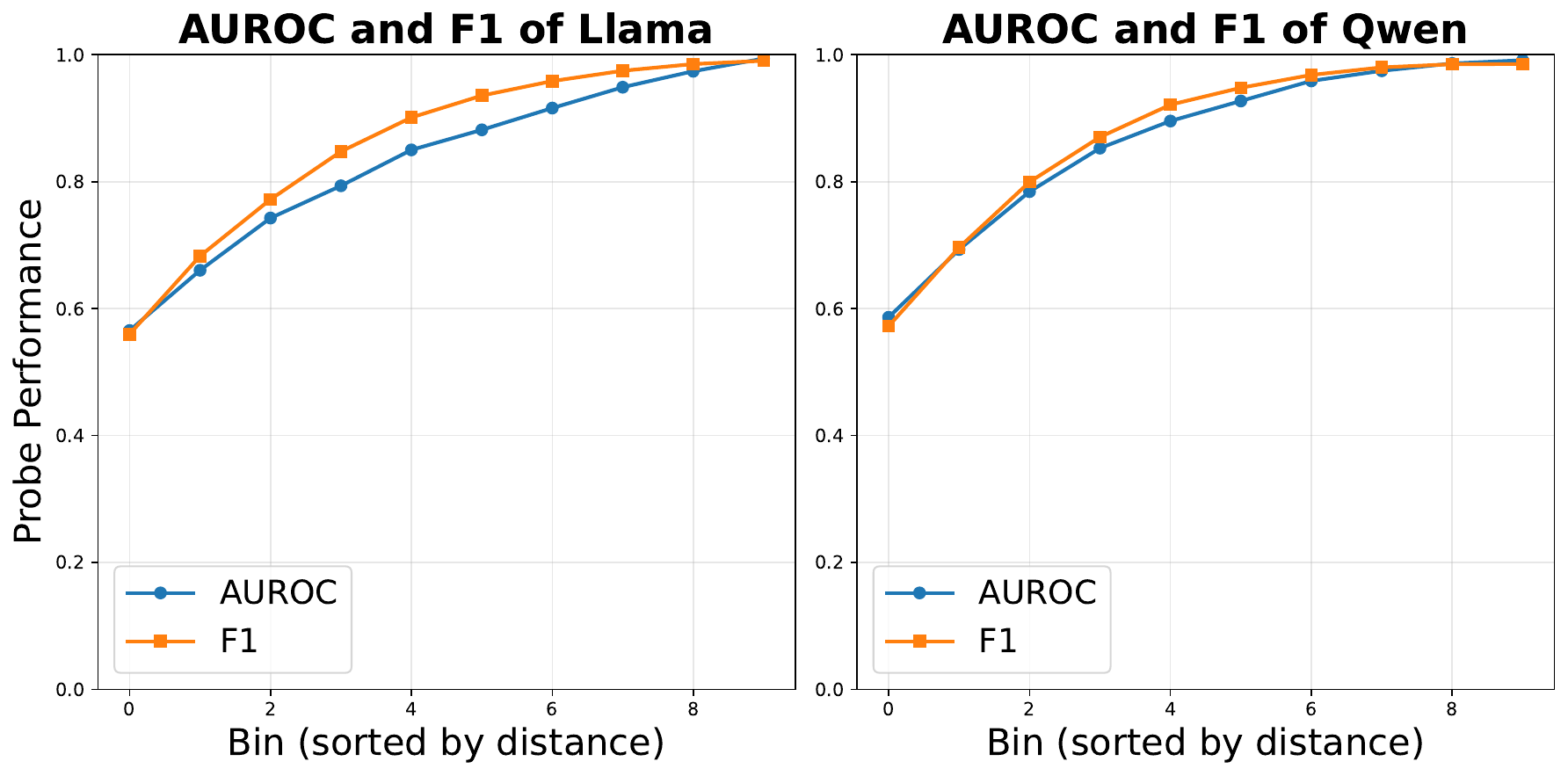}
    \caption{Probing performance vs. distance on TriviaQA. Bin0 denotes the nearest subset, and bin9 the farthest subset. Accuracy and F1 score increase as distance increases.}
    \label{fig:bin_auroc}
\end{figure}

To validate our hypothesis that the geometric distance to the probing hyperplane serves as a reliable proxy for sample quality, we partitioned the training data into three tiers: Ours-Farthest (75\%-100\%), Ours-Middle (37.5\%-62.5\%), and Ours-Nearest (0-25\%). As shown in Table~\ref{tab:ablation_distance}, a clear performance gradient is observed across all metrics. Specifically, on Qwen3-8B (TriviaQA), Ours-Farthest achieves an $\text{F1}_\text{rel}$ of 77.6, markedly surpassing the Middle (74.6) and Nearest (73.2) tiers. This degradation stems from the high label noise near the decision boundary: samples located near the hyperplane (the ``grey zone’’) represent instances where the model’s internal representations for known and unknown knowledge are highly entangled. Fine-tuning on these ambiguous samples introduces contradictory gradients that blur the model's knowledge boundaries. By selectively training on the ``farthest'' samples, \Ours effectively performs geometric denoising, ensuring that the model learns from high-confidence, linearly separable signals. 
To further illustrate this distance-quality correlation, we bin the training data into ten equal subsets sorted by distance in ascending order. As visualized in Figure~\ref{fig:bin_auroc}, the probe's predictive accuracy scales monotonically with distance. Notably, the AUROC for samples nearest to the hyperplane drops below 0.6, approaching random chance, confirming that proximity to the boundary is a primary source of aleatoric noise. In contrast, distal samples provide a clean signal for knowledge boundaries.
\begin{table*}[!htbp]
  \centering
  \input{table/ablation_x}
  \caption{Ablation study on the selection threshold $X$ (percentage of samples retained by geometric distance) on Qwen3-8B. The optimal setting $X{=}25\%$ balances data quantity and label noise. Lower $X$ causes underfitting due to insufficient data; higher $X$ reintroduces boundary noise.}
  \label{tab:ablation_x}
\end{table*}

\begin{table}

  \begin{center} 
  \input{table/ablation_distance}
  \end{center}
  \caption{Ablation study on distance-based data partitioning for abstention fine-tuning. Samples are partitioned into three tiers based on their geometric distance $|d(x)|$ to the probing hyperplane.}
\label{tab:ablation_distance}
\end{table}

We further compare this geometric denoising approach with R-Tuning-01, which conceptually shares a similar objective by training only on samples with 100\% response consistency (either all correct or all incorrect) across 10 independent sampling runs. While R-Tuning-01 attempts to filter ambiguity via external output stability, it remains inferior to our method.

\paragraph{Sensitivity to the Selection Threshold $X$}
We examine how sensitive \Ours is to the threshold $X$, which controls the fraction of samples retained by geometric distance. Table~\ref{tab:ablation_x} reports results on Qwen3-8B across five settings. Performance peaks at $X{=}25\%$ on the majority of metrics: on TriviaQA, this yields $\text{F1}_{\text{rel}}=77.6$, while on NQ it yields $\text{F1}_{\text{rel}}=61.6$. When $X$ is too small (e.g., $6.25\%$), performance drops due to insufficient training data; when $X$ is too large (e.g., $75\%$), noisy boundary samples are reintroduced, degrading reliability. These results confirm that $X{=}25\%$ strikes the best balance, and that \Ours is robust across a reasonable range of $X$.

\subsection{Effects of Positive Proportions in Training}
\begin{figure}
    \centering
    \includegraphics[width=0.45\textwidth]{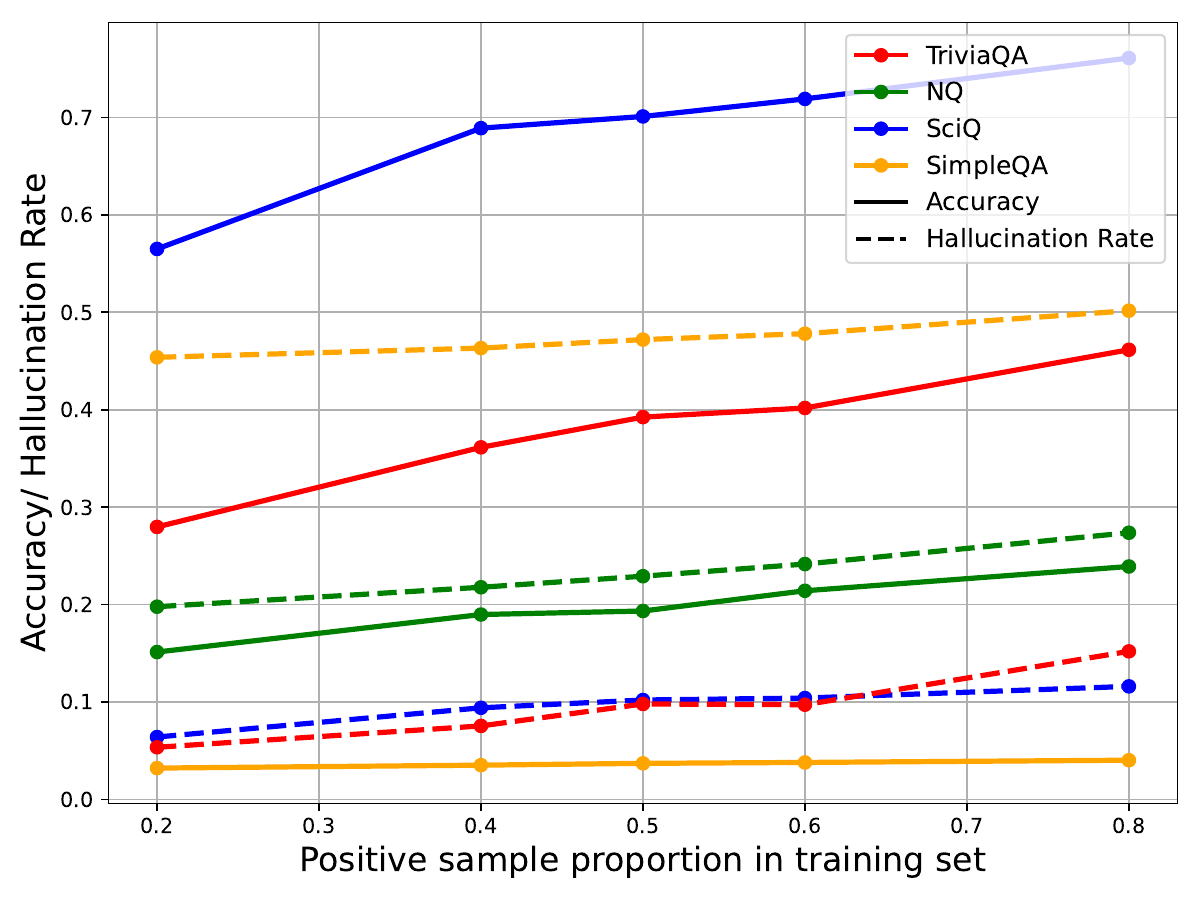}
    \caption{Accuracy and hallucination rate according to different positive proportions in the training set.}
    \label{fig:positive_ratio}
\end{figure}
We fix the training set size at 20,000 and conduct experiments under different positive sample proportions. Here, \textbf{positive samples} refer to question-answer pairs retaining the ground-truth answer (labeled ``known''), and \textbf{negative samples} refer to pairs whose target has been replaced with ``I don't know'' (labeled ``unknown''). \textbf{Accuracy} is the fraction of questions on which the model is willing to answer \emph{and} answers correctly; \textbf{hallucination rate} is the fraction on which the model answers but answers incorrectly. As shown in~\Cref{fig:positive_ratio}, both accuracy and hallucination rate increase with the proportion of positive samples in the training set. High-accuracy datasets like TriviaQA and SciQ exhibit significant sensitivity to changes in the positive-to-negative ratio within the training set, whereas NQ and SimpleQA show less pronounced variation. This suggests that excessively high negative sample ratios may cause models to over-abstain from known questions. The results also imply that abstention fine-tuning cannot eliminate over-abstention and hallucination simultaneously.

\subsection{Evaluation on Unanswerable Datasets}

\paragraph{Experimental Setup} To evaluate the model's ability for identifying unanswerable queries, we test performance on three specialized benchmarks: (1) \textbf{Alcuna}~\citep{ALCUNA}, containing synthetic entity-based questions; (2) \textbf{FalseQA}~\citep{FalseQA}, containing questions contradicting common sense; and (3) \textbf{Self-Aware (SA)}~\citep{self-aware}, containing inherently unanswerable questions.

\paragraph{Performance Analysis}

\begin{table}[htbp] 
\centering 
\input{table/unanswerable}
\caption{Abstention rates (\%) across specialized benchmarks. High scores indicate effective identification of unanswerable or deceptive queries.} 
\label{tab:unanswerable} 
\end{table}

As presented in Table~\ref{tab:unanswerable}, \Ours demonstrates highly competitive abstention rates, particularly with the TBG variant. While performance varies across benchmarks, our method shows notable strengths in specific scenarios. For instance, on the Alcuna dataset using Qwen3-8B, \Ours (TBG) achieves an abstention rate of 94.5\%, a substantial improvement over the 78.2\% reached by R-Tuning. Notably, overall rejection rates are higher on Alcuna than on FalseQA and SA. We attribute this to the nature of entities: while FalseQA and SA involve real-world concepts, Alcuna uses synthetic ones. This suggests that familiar domains trigger an illusion of knowledge, where the presence of known entities induces generative overconfidence. Consequently, models struggle more to recognize their epistemic limits in familiar contexts than in entirely novel, synthetic ones.

\section{Conclusion}
In this work, we introduce \Ours, a novel framework for abstention fine-tuning that leverages the latent geometry of LLMs. Moving beyond traditional methods that rely solely on external response accuracy, we propose a diagnostic perspective by analyzing the model's internal representation space. We identify a critical ``grey zone'' near the latent decision hyperplane, where ambiguous internal beliefs introduce significant noise that hinders a model's ability to perceive its own knowledge boundaries. By employing geometric denoising, \Ours systematically purges these ambiguous boundary samples, ensuring fine-tuning on high-fidelity, linearly separable signals. Extensive experiments across multiple models (Llama3, Qwen3) and benchmarks demonstrate that our approach significantly enhances model reliability and exhibits superior generalization in OOD, RAG, and deceptive scenarios.
% Ultimately, our findings suggest that aligning fine-tuning data with internal belief states is a more robust strategy for mitigating hallucinations and establishing precise knowledge boundaries than surface-level consistency heuristics.

\section*{Limitations}
\Ours has been validated on models ranging from 1.7B to 8B parameters (see Appendix~\ref{appendix:model_sizes}), but has not yet been evaluated at larger scales (e.g., 70B). Beyond scale, our method relies on linear probes to model truthfulness, which may not fully capture the complexity of truth-related representations in larger models; recent evidence suggests that a multi-dimensional framework may be necessary~\citep{yu2025directionsconesexploringmultidimensional}. Our experiments are also limited to short-form QA and RAG settings, leaving reasoning tasks and long-form generation as important directions for future work. Finally, while \Ours selects high-confidence training samples to reduce label noise, abstention fine-tuning inherently risks over-abstention—causing the model to refuse questions it could previously answer correctly. Addressing this trade-off, perhaps by incorporating uncertainty-aware objectives during pre-training, remains an open challenge.

\section*{Acknowledgement}
We sincerely thank all the reviewers for their feedback on the paper. 
This study is funded by Shenzhen Science and Technology Program (No. JCYJ20240813094612017) and Guangdong Province ZJRC Program (No. 2024QN11X145).

\section*{Ethics Statement}
In this research, we used publicly available datasets and we did not collect any personal information. We used AI for coding and paper polishing. Our method aims to improve the reliability of large language models through abstention. However, during deployment, our method may lead to over-abstention and cannot completely prevent hallucinations; therefore, caution should be exercised when using it in practice.
\bibliography{custom}

\appendix

% \section{Hyperparameter}
% \label{hyperparameter}

% This is an appendix.
\section{Scalability Across Model Sizes}
\label{appendix:model_sizes}
\begin{table*}[!htbp]
  \centering
  \input{table/model_sizes}
  \caption{Performance of R-Tuning and \Ours across different Qwen3 model sizes. All results are multiplied by 100. The best result per model size is bolded.}
  \label{tab:model_sizes}
\end{table*}
To verify that \Ours generalizes across models of varying capacity, we conduct experiments on Qwen3-1.7B and Qwen3-4B in addition to the main Qwen3-8B results. As shown in Table~\ref{tab:model_sizes}, \Ours consistently outperforms R-Tuning in overall reliability ($\text{F1}_{\text{rel}}$) across both model sizes and all four benchmarks. For example, on Qwen3-4B, \Ours achieves $\text{F1}_{\text{rel}}=76.9$ on TriviaQA and $60.2$ on NQ, compared to $74.6$ and $54.5$ for R-Tuning. On Qwen3-1.7B, the gains are similarly consistent. These results demonstrate that the geometric denoising principle of \Ours is model-size agnostic: the latent truthfulness hyperplane provides a reliable confidence signal regardless of parameter count.

\section{Dataset Details}
Table~\ref{tab:dataset_details} shows statistical information about the datasets, where SimpleQA and NQ are more challenging datasets, while TriviaQA and NQ are relatively simpler. Alcuna, FalseQA and Self-Aware contain unanswerable questions.
\label{dataset_detail}
\begin{table}[htbp!]
  \centering

\input{table/dataset_details.tex}
 \caption{Dataset details. Llama (Llama3-8B-Instruct) and Qwen (Qwen3-8B) are the initial models used in experiments. Acc. (Accuracy) is for reference.}
\label{tab:dataset_details}
\end{table}

\section{Prompts}

\input{appendix/prompts}

\section{Judge Robustness and Abstention Format}
\label{appendix:judge_robustness}
\input{appendix/judge_robustness}

\section{Case Study}
\label{appendix:case_study}
\input{appendix/case_study}

\end{document}

%% file: table/algotithm.tex
\begin{algorithm}
\caption{\Ours TBG Process}\label{alg:cap}
\begin{algorithmic}[1]
\Require Initial model $M$, QA dataset $D_{\text{src}}=D0+D1$, percentage threshold $X$
\Ensure Fine-tuned model $M_{\text{fine-tuned}}$
\State \textbf{Step 1: Identify Knowledge Boundary}
\State \quad Test $M$ on $D0$
\State \quad Split $D0$ into $D0_{\text{ik}}$  and $D0_{\text{idk}}$ by accuracy
\State \textbf{Step 2: Curate Subsets Based on Distance}
\State \quad Get representation $x=f_\text{LLM}(q)$, $q\in D0$ % $(q,a)\in D0_\text{ik}\cup D0_\text{idk}$
\State \quad Train linear probe on $D0$: $f_\text{probe}(x) = \sigma(w^\top x + b)$
\State \quad Compute distance on $D1$: $d(x) = \frac{| w^\top x + b |}{||w||_2}$

% \State \quad Sort confidence scores in descending order
\State \quad Determine threshold $\theta$ as the $X\%$-th quantile of sorted $d(x)$
\State \quad Select top $X\%$ samples: $D1^{\text{selected}} \gets \{ x \mid |d(x)| > \theta \}$
\State \textbf{Step 3: Abstention Fine-tuning}
\State \quad Split $D1^\text{selected}$ into $D1_{\text{ik}}^{\text{selected}}$  ($d(x)>0$) and $D1_{\text{idk}}^{\text{selected}}$ ($d(x)<0$).
\State \quad Replace ground truth of  $D1^{\text{selected}}_{\text{idk}}$ with ``I don't know.''

\State \quad Fine-tune $M$ with cross-entropy loss on $D1^{\text{selected}}$
% ${\cal L}_{CE}(p_\theta) = - \sum_{q \in Q} \sum_{t=1}^{|y^{(q)}|} \log p_\theta \bigl( y^{(q)}_t \mid \text{I}, q, y^{(q)}_{t-1} \bigr)$
\State \quad Return $M_{\text{fine-tuned}}$
\end{algorithmic}
\end{algorithm}

%% file: table/Abstention_confusion_matrix.tex
\begin{tabular}{l|c|c|c}
    \toprule
    \diagbox{GT}{R} & \makecell{Correctly \\ answered} & \makecell{Wrongly \\ answered} & Abstained \\
    \midrule
    Known  & $N_1$ & $N_2$ & $N_3$ \\
    Unknown &  --   & $N_4$ & $N_5$ \\
    \bottomrule
\end{tabular}

%% file: table/main_reults.tex
% \usepackage{booktabs}
% \usepackage[normalem]{ulem}
\setlength{\tabcolsep}{3pt}
\begin{tabular}{@{}lllllllllllll@{}}
\toprule

\textbf{Dataset} &
  \multicolumn{3}{c}{\textbf{TriviaQA}} &
  \multicolumn{3}{c}{\textbf{NQ}} &
  \multicolumn{3}{c}{\textbf{SciQ}} &
  \multicolumn{3}{c}{\textbf{SimpleQA}} \\ \midrule
\textbf{Method} &
  $\text{F1}_{\text{ans}}$ &
  $\text{F1}_{\text{abs}}$ &
  $\text{F1}_{\text{rel}}$ &
  $\text{F1}_{\text{ans}}$ &
  $\text{F1}_{\text{abs}}$ &
  $\text{F1}_{\text{rel}}$ &
  $\text{F1}_{\text{ans}}$ &
  $\text{F1}_{\text{abs}}$ &
  $\text{F1}_{\text{rel}}$ &
  $\text{F1}_{\text{ans}}$ &
  $\text{F1}_{\text{abs}}$ &
  $\text{F1}_{\text{rel}}$ \\ \midrule
\multicolumn{13}{c}{\textbf{Llama3-8B-Instruct}} \\ \midrule
\textbf{IDK} &
  78.7 &
  35.3 &
  48.8 &
  \textbf{61.4} &
  56.3 &
  58.7 &
  \textbf{83.3} &
  5.6 &
  10.4 &
  14.6 &
  64.0 &
  23.8 \\
\textbf{Uncertainty} &
  67.7 &
  46.4 &
  55.0 &
  45.9 &
  18.0 &
  25.9 &
  64.7 &
  17.9 &
  28.0 &
  10.0 &
  52.2 &
  16.8 \\
\textbf{R-Tuning} &
  77.3 &
  71.6 &
  74.4 &
  47.0 &
  78.1 &
  58.7 &
  69.9 &
  58.2 &
  63.5 &
  14.6 &
  \textbf{96.1} &
  25.4 \\
\textbf{Probe-Tuning TBG} &
  78.8 &
  72.8 &
  75.7 &
  51.6 &
  {\ul 79.1} &
  62.5 &
  81.4 &
  53.4 &
  64.5 &
  12.8 &
  \textbf{96.1} &
  22.6 \\
\rowcolor[HTML]{ECF4FF} 
\textbf{\Ours TBG} &
  \textbf{80.9} &
  \textbf{73.7} &
  \textbf{77.1} &
  {\ul 54.0} &
  78.4 &
  \textbf{64.0} &
  81.9 &
  58.5 &
  {\ul 68.3} &
  {\ul 18.4} &
  94.8 &
  30.7 \\
\textbf{Probe-Tuning SLT} &
  75.8 &
  71.3 &
  73.4 &
  44.8 &
  78.3 &
  56.9 &
  75.4 &
  {\ul 58.9} &
  66.1 &
  {\ul 18.4} &
  95.8 &
  {\ul 30.9} \\
\rowcolor[HTML]{ECF4FF} 
\textbf{\Ours SLT} &
  {\ul 79.8} &
  \textbf{73.7} &
  {\ul 76.7} &
  52.6 &
  \textbf{79.4} &
  {\ul 63.3} &
  {\ul 82.4} &
  \textbf{59.8} &
  \textbf{69.3} &
  \textbf{18.6} &
  96.0 &
  \textbf{31.2} \\ \midrule
\multicolumn{13}{c}{\textbf{Qwen3-8B}} \\ \midrule
\textbf{IDK} &
  74.0 &
  55.1 &
  63.2 &
  {\ul 55.1} &
  65.4 &
  59.8 &
  81.8 &
  44.5 &
  57.6 &
  11.7 &
  58.6 &
  19.5 \\
\textbf{Uncertainty} &
  {\ul 75.9} &
  38.8 &
  51.3 &
  \textbf{57.3} &
  52.1 &
  54.6 &
  70.3 &
  38.6 &
  49.8 &
  12.2 &
  {\ul 66.6} &
  20.6 \\
\textbf{R-Tuning} &
  75.8 &
  74.3 &
  75.0 &
  50.3 &
  70.5 &
  58.7 &
  \textbf{86.5} &
  45.1 &
  {\ul 59.3} &
  12.6 &
  63.6 &
  21.0 \\
\textbf{Probe-Tuning TBG} &
  75.0 &
  71.7 &
  73.3 &
  51.2 &
  70.5 &
  59.4 &
  \textbf{86.5} &
  44.2 &
  58.5 &
  12.6 &
  63.4 &
  21.0 \\
\rowcolor[HTML]{ECF4FF} 
\textbf{\Ours TBG} &
  \textbf{77.7} &
  \textbf{77.4} &
  \textbf{77.6} &
  54.3 &
  {\ul 71.3} &
  \textbf{61.6} &
  {81.8} &
  {\ul 47.2} &
  59.8 &
  {\ul 13.0} &
  \textbf{67.3} &
  {\ul 21.7} \\
\textbf{Probe-Tuning SLT} &
  75.4 &
  72.8 &
  74.1 &
  50.1 &
  69.2 &
  58.1 &
  85.7 &
  45.8 &
  {\ul 60.0} &
  12.3 &
  62.0 &
  20.6 \\
\rowcolor[HTML]{ECF4FF} 
\textbf{\Ours SLT} &
  75.6 &
  {\ul 75.6} &
  {\ul 75.6} &
  51.6 &
  \textbf{71.7} &
  {\ul 60.0} &
  {84.6} &
  \textbf{48.5} &
   \textbf{61.6}&
  \textbf{13.8} &
  65.9 &
  \textbf{22.8} \\ \bottomrule
\end{tabular}

%% file: table/rag.tex
 \begin{tabular}{@{}lllllllllll@{}}
\toprule
                   & \multicolumn{5}{c}{\textbf{Golden}}      & \multicolumn{5}{c}{\textbf{Golden \& RRN}} \\ \cmidrule(l){2-11} 
\multirow{-2}{*}{} &  $\text{F1}_{\text{ans}}$ &  $\text{F1}_{\text{abs}}$ &  $\text{F1}_{\text{rel}}$ & Acc. & Hallu. &  $\text{F1}_{\text{ans}}$  &  $\text{F1}_{\text{abs}}$  &  $\text{F1}_{\text{rel}}$ & Acc. & Hallu. \\ \midrule
ICL                & -      & -      & -      & 71.2 & 28.8   & -       & -       & -      & 63.8 & 36.2   \\
IDK                & 76.1   & 45.5   & 57.0   & 58.1 & 17.9   & 71.6    & 45.6    & 55.7   & 52.7 & 23.6   \\
R-Tuning           & 80.8   & 52.5   & 63.7   & 61.0 & 11.8   & 74.4    & 53.3    & 62.1   & 50.4 & 15.3   \\
Probe-tuning TBG   & 83.8   & 47.9   & 61.0   & 67.3 & 13.0   & 81.1    & 48.6    & 60.8   & 63.3 & 16.8   \\
\rowcolor[HTML]{ECF4FF} 
\Ours TBG           & 75.2   & 52.5   & 61.8   & 51.5 & 9.2    & 74.3    & 56.9    & 64.5   & 49.1 & 11.4   \\
Probe-Tuning SLT   & 79.1   & 52.7   & 63.2   & 54.8 & 11.9   & 75.9    & 54.7    & 63.5   & 50.0 & 15.7   \\
\rowcolor[HTML]{ECF4FF} 
\Ours SLT           & 79.9   & 61.1   & 69.3   & 55.8 & 9.2    & 75.8    & 57.9    & 65.7   & 53.4 & 13.0   \\ \bottomrule
\end{tabular}

%% file: table/ablation_x.tex
\setlength{\tabcolsep}{3pt}
\begin{tabular}{@{}lllllllllllll@{}}
\toprule
\textbf{Dataset} &
  \multicolumn{3}{c}{\textbf{TriviaQA}} &
  \multicolumn{3}{c}{\textbf{NQ}} &
  \multicolumn{3}{c}{\textbf{SciQ}} &
  \multicolumn{3}{c}{\textbf{SimpleQA}} \\ \midrule
\textbf{$X$} &
  $\text{F1}_{\text{ans}}$ &
  $\text{F1}_{\text{abs}}$ &
  $\text{F1}_{\text{rel}}$ &
  $\text{F1}_{\text{ans}}$ &
  $\text{F1}_{\text{abs}}$ &
  $\text{F1}_{\text{rel}}$ &
  $\text{F1}_{\text{ans}}$ &
  $\text{F1}_{\text{abs}}$ &
  $\text{F1}_{\text{rel}}$ &
  $\text{F1}_{\text{ans}}$ &
  $\text{F1}_{\text{abs}}$ &
  $\text{F1}_{\text{rel}}$ \\ \midrule
6.25\%  & 66.9 & 69.2 & 68.1 & 42.2 & 70.0 & 52.6 & 71.0 & 40.5 & 51.6 & 11.2 & 63.8 & 19.1 \\
12.5\%  & 72.8 & 74.0 & 73.4 & 48.2 & 70.9 & 57.4 & \textbf{85.6} & 48.5 & \textbf{61.9} & 12.1 & 64.8 & 20.4 \\
\rowcolor[HTML]{ECF4FF}
25\%    & \textbf{77.7} & \textbf{77.4} & \textbf{77.6} & \textbf{54.3} & \textbf{71.3} & \textbf{61.6} & 81.8 & 47.2 & 59.8 & \textbf{13.0} & \textbf{67.3} & \textbf{21.7} \\
50\%    & 75.6 & 75.6 & 75.6 & 51.1 & 70.9 & 59.4 & 85.4 & 45.7 & 59.6 & 12.5 & 64.1 & 20.9 \\
75\%    & 75.8 & 75.0 & 75.4 & 50.7 & 70.4 & 58.9 & 86.0 & \textbf{45.9} & 59.9 & 12.3 & 63.3 & 20.6 \\ \bottomrule
\end{tabular}

%% file: table/ablation_distance.tex
\setlength{\tabcolsep}{1.5pt} 
\begin{tabular}{@{}l|lll|lll@{}}
\toprule
Dataset                  & \multicolumn{3}{c|}{TriviaQA} & \multicolumn{3}{c}{NQ}          \\ \midrule
Metric                   & $\text{F1}_{\text{ans}}$   & $\text{F1}_{\text{abs}}$   & $\text{F1}_{\text{rel}}$  & $\text{F1}_{\text{ans}}$ & $\text{F1}_{\text{abs}}$        & $\text{F1}_{\text{rel}}$ \\ \midrule
Ours TBG & \textbf{77.7} & \textbf{77.4} & \textbf{77.6} & \textbf{54.3} & 71.3 & \textbf{61.6} \\
-middle              & 74.9     & 74.2     & 74.6    & 50.0   & 70.2          & 58.4   \\
-nearest             & 75.2     & 71.2     & 73.2    & 49.7   & 69.6          & 58.0   \\ \midrule
Ours SLT & 75.6     & 75.6     & 75.6    & 51.6   & 71.7          & 60.0   \\
-middle             & 75.4     & 72.8     & 74.1    & 50.1   & 69.2          & 58.1   \\
-nearest            & 73.8     & 67.5     & 70.5    & 48.9   & 67.2          & 56.6   \\ \midrule
R-Tuning-01              & 75.5     & 77.0     & 76.2    & 51.6   & \textbf{71.8} & 60.0   \\ \bottomrule
\end{tabular}

%% file: table/unanswerable.tex
\begin{tabular}{@{}llll@{}}
\toprule
\textbf{Method}                   & \textbf{Alcuna} & \textbf{FalseQA} & \textbf{SA}   \\ \midrule
\multicolumn{4}{c}{\textbf{Llama3-8b-Instruction}}                                     \\ \midrule
\textbf{IDK}                      & 78.4            & 49.8             & 50.9          \\
\textbf{Uncertainty}              & 19.2            & 8.5              & 15.4          \\
\textbf{R-Tuning}                 & 98.3            & 91.8             & 98.1          \\
\textbf{Probe-Tuning TBG}         & 99.2            & 91.8             & \textbf{99.7} \\
\textbf{\Ours TBG} & 98.2            & 92.4             & 99.2          \\
\textbf{Probe-Tuning SLT}         & \textbf{99.7}   & \textbf{95.3}    & \textbf{99.7} \\
\textbf{\Ours SLT} & 99.0            & 90.6             & 98.7          \\ \midrule
\multicolumn{4}{c}{\textbf{Qwen3-8B}}                                                  \\ \midrule
\textbf{IDK}                      & 85.9            & 62.4             & 74.4          \\
\textbf{Uncertainty}              & 38.5            & 24.5             & 39.6          \\
\textbf{R-Tuning}                 & 78.2            & 67.5             & 82.6          \\
\textbf{Probe-Tuning TBG}         & 84.0            & \textbf{72.2}    & 84.1          \\
\textbf{\Ours TBG} & \textbf{94.5}   & 67.6             & \textbf{88.3} \\
\textbf{Probe-Tuning SLT}         & 69.7            & 65.2             & 80.6          \\
\textbf{\Ours SLT} & 94.3            & 70.5             & 86.9          \\ \bottomrule
\end{tabular}

%% file: table/model_sizes.tex
\setlength{\tabcolsep}{3pt}
\begin{tabular}{@{}lllllllllllll@{}}
\toprule
\textbf{Dataset} &
  \multicolumn{3}{c}{\textbf{TriviaQA}} &
  \multicolumn{3}{c}{\textbf{NQ}} &
  \multicolumn{3}{c}{\textbf{SciQ}} &
  \multicolumn{3}{c}{\textbf{SimpleQA}} \\ \midrule
\textbf{Method} &
  $\text{F1}_{\text{ans}}$ &
  $\text{F1}_{\text{abs}}$ &
  $\text{F1}_{\text{rel}}$ &
  $\text{F1}_{\text{ans}}$ &
  $\text{F1}_{\text{abs}}$ &
  $\text{F1}_{\text{rel}}$ &
  $\text{F1}_{\text{ans}}$ &
  $\text{F1}_{\text{abs}}$ &
  $\text{F1}_{\text{rel}}$ &
  $\text{F1}_{\text{ans}}$ &
  $\text{F1}_{\text{abs}}$ &
  $\text{F1}_{\text{rel}}$ \\ \midrule
\multicolumn{13}{c}{\textbf{Qwen3-1.7B}} \\ \midrule
R-Tuning &
  56.7 & 87.0 & 68.7 &
  \textbf{38.8} & 89.4 & \textbf{54.1} &
  66.5 & \textbf{73.6} & 69.9 &
  5.9 & 63.3 & 10.8 \\
\rowcolor[HTML]{ECF4FF}
\textbf{\Ours} &
  \textbf{56.9} & \textbf{89.0} & \textbf{69.5} &
  38.0 & \textbf{92.2} & 53.8 &
  \textbf{77.0} & 73.2 & \textbf{75.0} &
  \textbf{6.8} & \textbf{98.5} & \textbf{12.7} \\ \midrule
\multicolumn{13}{c}{\textbf{Qwen3-4B}} \\ \midrule
R-Tuning &
  69.4 & 80.6 & 74.6 &
  44.1 & 71.2 & 54.5 &
  83.0 & \textbf{63.3} & 71.8 &
  10.9 & 65.2 & 18.7 \\
\rowcolor[HTML]{ECF4FF}
\textbf{\Ours} &
  \textbf{71.7} & \textbf{82.9} & \textbf{76.9} &
  \textbf{49.1} & \textbf{77.6} & \textbf{60.2} &
  \textbf{85.1} & 62.6 & \textbf{72.1} &
  \textbf{11.3} & \textbf{72.6} & \textbf{19.5} \\ \bottomrule
\end{tabular}

%% file: table/dataset_details.tex
% \begin{tabular}{@{}llll@{}}
% \toprule
% \textbf{Test Dataset} & \textbf{Size} & \textbf{\begin{tabular}[c]{@{}l@{}}Llama\\  Acc.\end{tabular}} & \textbf{\begin{tabular}[c]{@{}l@{}}Qwen \\ Acc.\end{tabular}} \\ \midrule
% \textbf{TriviaQA (ID)} & 11313 & 65.55 & 54.65 \\
% \textbf{NQ (OOD)}       & 3610  & 40.86 & 34.04 \\
% \textbf{SciQ (OOD)}     & 1000  & 72.50 & 81.60 \\
% \textbf{SimpleQA (OOD)} & 4326  & 6.80  & 5.59  \\ \bottomrule
% \end{tabular}

% \usepackage{booktabs}

\begin{tabular}{@{}llll@{}}
\toprule
 & \textbf{Size} & \textbf{\begin{tabular}[c]{@{}l@{}}Llama \\ Acc.\end{tabular}} & \textbf{\begin{tabular}[c]{@{}l@{}}Qwen \\ Acc.\end{tabular}} \\ \midrule
\textbf{TriviaQA Train} & 87622 & 66.17 & 54.93 \\
\textbf{TriviaQA Val}   & 11313 & 65.55 & 54.65 \\
\textbf{NQ}             & 3610  & 40.86 & 34.04 \\
\textbf{SciQ}           & 1000  & 72.50 & 81.60 \\
\textbf{SimpleQA}       & 4326  & 6.80  & 5.59  \\
\textbf{Alcuna}         & 2001  & -     & -     \\
\textbf{FalseQA}        & 1374  & -     & -     \\
\textbf{Self-Aware}     & 1032  & -     & -     \\ \bottomrule
\end{tabular}

%% file: appendix/prompts.tex
During training, we use the following instruction: 

\textit{You are a helpful and truthful AI assistant. You should answer the question as briefly as possible, if you don't know, please just say 'I don't know."}.

We use a 6-shot prompt for evaluation across all methods. For the ICL baseline, the prompt consists of six examples of direct answering. In contrast, prompts for abstention-aware methods include a balanced mix of three answering examples and three abstention examples.
\definecolor{almond}{RGB}{239, 222, 205}

    \begin{tcolorbox}[ title=\textbf{ICL Prompt}]
    \label{temp:icl}
        
        Answer the following questions as briefly as possible.\\
        \\
        Question: {\texttt{\{demo question 1\}}}\\
        Answer: {\texttt{\{demo answer 1\}}}\\
        \\
        Question: {\texttt{\{demo question 2\}}}\\
        Answer: {\texttt{\{demo answer 2\}}}\\
        \\
        ...\\
        \\
        Question: {\texttt{\{input question\}}}\\
        Answer: \\
    \end{tcolorbox}
    
    \begin{tcolorbox}[title=\textbf{Abstention-aware Prompt}]
    \label{temp:icl-idk}
        Answer the following questions as briefly as possible. If you don't know the answer, please simply say ``I don't know.''\\
        \\
        Question: {\texttt{\{demo question 1\}}}\\
        Answer: {\texttt{\{demo answer 1\}}}\\
        \\
        Question: {\texttt{\{demo question 2\}}}\\
        Answer: {\texttt{I don't know.}}\\
        \\
        ...\\
        \\
        Question: {\texttt{\{input question\}}}\\
        Answer: 
    \end{tcolorbox}
    
    \begin{tcolorbox}[ title=\textbf{Uncertainty Prompt}]
    \label{temp:icl-cot}
        You should answer the question as briefly as possible, then present your confidence. If you are sure about your answer, please say ``I am sure'' after your answer; otherwise, say ``I am unsure''.\\
        \\
        Question: {\texttt{\{demo question 1\}}}\\
        Answer: {\texttt{\{demo answer 1\}}} I am sure.\\
        \\
        Question: {\texttt{\{demo question 2\}}}\\
        Answer: {\texttt{\{demo answer 2\}}} I am unsure.\\
        \\
        ...\\
        \\
        Question: {\texttt{\{input\_question\}}}\\
        Answer: 
    \end{tcolorbox}

    \begin{tcolorbox}[ title=\textbf{LLM judge Prompt}]
    \label{temp:illmjudge}
        We are assessing the quality of answers to the following question: \textit{input question}\\
        The following are expected answers to this question: {\textit{input ground truth}}\\
        The proposed answer is \textit{proposed answer} \\
        Within the context of the question, does the proposed answer mean the same as the expected answer?\\
        Respond only with yes or no.\\
        Here are some examples:
        \\
        Question: \textit{demo question 1}  \\
        Expected answer: \textit{demo ground truth 1}\\
        Proposed answer: \textit{demo correct answer 1} \\
        Response: \textit{yes}\\ \\ 
        Question: \textit{demo question 2}  \\
        Expected answer: \textit{demo ground truth 2}\\
        Proposed answer: \textit{demo wrong answer 2} \\
        Response: \textit{no}\\\\
        ...
        \\\\
    Now evaluate the following:\\
    Question: \textit{input question}\\
    Expected answer: \textit{input ground truth}\\
    Proposed answer: \textit{input proposed answer}\\
    Response:
    \end{tcolorbox}

%% file: appendix/judge_robustness.tex
Correctness in our experiments is determined via Llama3.1-8B-Instruct with a 6-shot judging prompt, which may introduce biases such as judge errors and phrasing sensitivity. To assess this, we performed cross-validation by running the same evaluation with a second judge, Gemma2-9B-IT, and computed Cohen's Kappa between the two judges across all four datasets.

\begin{table}[h]
  \centering
  \input{table/judge_kappa}
  \caption{Inter-judge agreement (Cohen's Kappa) between Llama3.1-8B-Instruct and Gemma2-9B-IT across evaluation datasets.}
  \label{tab:judge_kappa}
\end{table}

As shown in Table~\ref{tab:judge_kappa}, agreement is high on TriviaQA ($\kappa=0.956$), NQ ($\kappa=0.912$), and SimpleQA ($\kappa=0.932$). SciQ shows slightly lower agreement ($\kappa=0.828$); manual inspection reveals that some SciQ questions admit multiple valid answers while the dataset provides only a single reference label, causing both judges to disagree on borderline-correct responses.

Regarding robustness to abstention format, since we fine-tune with a fixed ``I don't know'' label, the fine-tuned model outputs exclusively this phrase for abstentions, making format sensitivity negligible for the SFT model. For the initial (ICL) model, any refusal-style output is treated as an abstention, consistent with standard practice.

%% file: table/judge_kappa.tex
\begin{tabular}{@{}lc@{}}
\toprule
\textbf{Dataset} & \textbf{Cohen's Kappa} \\ \midrule
TriviaQA  & 0.9558 \\
NQ        & 0.9119 \\
SciQ      & 0.8281 \\
SimpleQA  & 0.9318 \\ \bottomrule
\end{tabular}

%% file: appendix/case_study.tex
\subsection{TBG vs.\ SLT Failure Mode Analysis}

To understand why \Ours can still hallucinate in the RAG setting, we analyze two typical failure cases using Qwen3-8B. As shown in~\Cref{tab:case_study_rag}, the LLM receives the question, golden retrieval, and relevant retrieval noise sequentially. Probe confidence is reported after each new input segment. Bold text marks the ground-truth answer; strikethrough text marks the model's incorrect output.

\begin{table*}[!htbp]
  \centering
  \input{table/case_study_rag}
  \caption{Hallucination cases in the Golden \& RRN RAG setting. Confidence is the probe's predicted probability that the model knows the answer (higher = more confident). The model hallucinates despite initially high confidence due to interference from retrieval noise.}
  \label{tab:case_study_rag}
\end{table*}

\textbf{Case 1 (confidence collapse then recovery).}
The question is about the origin of Latin. The model starts with moderate confidence (0.605). After reading the golden retrieval—which contains the correct answer (\textit{Italic languages}) but also many distracting spans—confidence drops below 0.5 (0.456), suggesting that irrelevant content within the golden passage overwhelms the model's reading comprehension. Exposure to the subsequent noise passage then raises confidence back to 0.721, and the model outputs an incorrect answer (\textit{River Tiber}) from the noisy context.

\textbf{Case 2 (confidence suppressed by noise).}
The question asks about Anakin Skywalker's identity. The model is highly confident on the question alone (0.983) and remains so after the golden passage (0.946), which correctly points to \textit{Darth Vader}. However, the noise passage introduces \textit{Jake Lloyd} as a related entity, dropping confidence to 0.591. Despite the drop, confidence stays above the abstention threshold, and the model outputs the noise-introduced wrong answer. These cases illustrate a core challenge: retrieval noise can either inflate or suppress probe confidence, causing the model to answer when it should abstain.

%% file: table/case_study_rag.tex
\setlength{\tabcolsep}{4pt}
\renewcommand{\arraystretch}{1.3}
\begin{tabular}{@{}p{0.09\linewidth}p{0.75\linewidth}c@{}}
\toprule
\textbf{Input} & \textbf{Content} & \textbf{Conf.} \\ \midrule
\multicolumn{3}{l}{\textit{Case 1 — GT: \textbf{Italic languages}; \ Model output: \sout{River Tiber}}} \\[2pt]
Question   & Where did the Latin language originate from? & 0.605 \\
Golden     & Latin is a member of the broad family of \textbf{Italic languages}. Its alphabet, the Latin alphabet, emerged from the Old Italic alphabets, which in turn were derived from the Greek and Phoenician scripts. Historical Latin came from the prehistoric language of the Latium region, specifically around the \sout{River Tiber}, where Roman civilization first developed. & 0.456 \\
Noise      & The solecisms and non-Classical usages occasionally found in late Latin texts also shed light on the spoken language. A windfall source lies in the chance finds of wax tablets such as those found at Vindolanda on Hadrian's Wall. The Romance languages, a major branch of the Indo-European language family, comprise all languages that descended from Latin, the language of the Roman Empire. & 0.721 \\ \midrule
\multicolumn{3}{l}{\textit{Case 2 — GT: \textbf{Darth Vader}; \ Model output: \sout{Jake Lloyd}}} \\[2pt]
Question   & Who was Anakin Skywalker? & 0.983 \\
Golden     & On May 12, 2000, Christensen announced that he would be starring as Anakin Skywalker in the prequel films. The casting director reviewed about 1,500 other candidates before director George Lucas selected Christensen, who needed an actor with ``that presence of the Dark Side.'' This was essential to solidify Anakin Skywalker's transformation into \textbf{Darth Vader}. & 0.946 \\
Noise      & As a result, he decided to no longer keep all owned ``Star Wars'' memorabilia. \sout{Jake Lloyd} (born March 5, 1989) is an American former actor who played young Anakin Skywalker in the 1999 film, the first in the ``Star Wars'' prequel trilogy. & 0.591 \\ \bottomrule
\end{tabular}